\documentclass[conference]{IEEEtran}
\IEEEoverridecommandlockouts
\usepackage{cite}
\usepackage{amsmath,amssymb,amsfonts}
\usepackage{algorithmic}
\usepackage{graphicx}
\usepackage{textcomp}
\usepackage{xcolor}
\usepackage{subfigure}
\usepackage{graphicx}
\usepackage{mathrsfs}

\usepackage{lineno}
\usepackage{times}
\usepackage{bm}
\usepackage{amsmath}
\usepackage{graphicx}
\usepackage{amsfonts}
\usepackage{enumerate}
\usepackage{array}
\usepackage{multirow}
\usepackage{algorithm}
\usepackage{algorithmic}
\usepackage{float}
\usepackage{color}
\usepackage{booktabs}
\usepackage{makecell}

\def\BibTeX{{\rm B\kern-.05em{\sc i\kern-.025em b}\kern-.08em
    T\kern-.1667em\lower.7ex\hbox{E}\kern-.125emX}}
\begin{document}
\title{Comprehensive Performance Evaluation of YOLOv11, YOLOv10, YOLOv9, YOLOv8  and YOLOv5 on  Object Detection of  Power Equipment\\
\thanks{This work was supported by the Key Natural Science Foundation of Higher Education Institutions of Anhui Province under grant 2024AH050154, the Open Project of Anhui Province Key Laboratory of Special and Heavy Load Robot under grant TZJQR005-2024, the Project of Anhui International Joint Research Center for Metallurgical Process and System Science, Anhui University of Technology under grant 2023002,the Open Project of Key Laboratory of Multidisciplinary Management and Control of Complex Systems of Anhui Higher Education Institutes, Anhui University of Technology under grant CS2023-ZD01.
}\thanks{Corresponding Author: Xihong Fei, Email: fxhong@mail.ustc.edu.cn}
}
\author{
\IEEEauthorblockN{Zijian He\IEEEauthorrefmark{1},  Kang Wang\IEEEauthorrefmark{1}, \IEEEmembership{Member,~IEEE}, Tian Fang\IEEEauthorrefmark{1}, Lei Su\IEEEauthorrefmark{1}, Rui Chen\IEEEauthorrefmark{2},  Xihong Fei\IEEEauthorrefmark{1}, \IEEEmembership{Member,~IEEE} } 
\IEEEauthorblockA{\IEEEauthorrefmark{1}School of Electrical and Information Engineering, Anhui University of Technology, Ma’anshan 243002, China} 
\IEEEauthorblockA{\IEEEauthorrefmark{2}School of Electrical and Optoelectronic Engineering, West Anhui University, Lu’an, 237012, China} 
}
\maketitle
%
%

\begin{abstract}

With the rapid development of global industrial production, the demand for reliability in power equipment has been continuously increasing. Ensuring the stability of power system operations requires accurate methods to detect potential faults in power equipment, thereby guaranteeing the normal supply of electrical energy. In this article, the performance of YOLOv5, YOLOv8, YOLOv9, YOLOv10, and the state-of-the-art YOLOv11 methods was comprehensively evaluated for power equipment object detection. Experimental results demonstrate that the mean average precision (mAP) on a public dataset for power equipment was 54.4\%, 55.5\%, 43.8\%, 48.0\%, and 57.2\%, respectively, with the YOLOv11 achieving the highest detection performance. Moreover, the YOLOv11 outperformed other methods in terms of recall rate and exhibited superior performance in reducing false detections. In conclusion, the findings indicate that the YOLOv11 model provides a reliable and effective solution for power equipment object detection, representing a promising approach to enhancing the operational reliability of power systems.

\end{abstract}


\section{Introduction}
\begin{figure*}
  \begin{center}
  \includegraphics[width=0.86\textwidth]{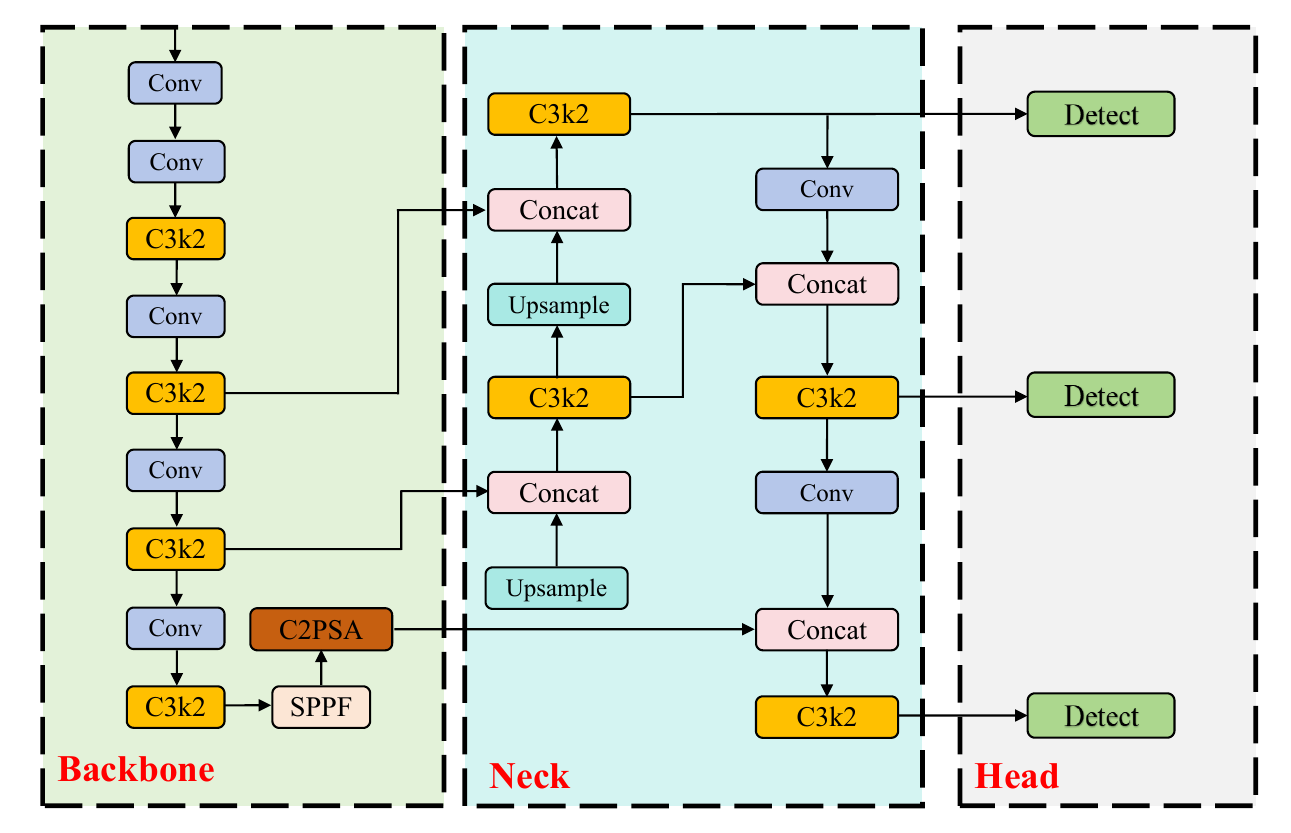}
  \setlength{\abovecaptionskip}{-1pt}
  \caption{Overall architecture of YOLOv11 network. }
  \label{fig1}
  \end{center}
\end{figure*}
In recent years, the rapid expansion of global industry has highlighted the importance of environmental protection. Reliance on burning conventional energy sources greatly increases greenhouse gas emissions and the release of harmful pollutants\cite{liu2024electrical,liu2023summary}.  Therefore, the demand for renewable and low-pollution energy sources has surged, aligning with the objectives of sustainable development\cite{mohtasham2015renewable}.  Renewable energy, in particular, offers a promising pathway to achieving these goals due to its inherent advantages of renewability and minimal environmental impact. The increasing adoption of renewable energy, especially for power generation, has heightened expectations for the stability and reliability of these systems. However, traditional manual detection methods fall short in meeting these demands. Such methods are not only inefficient but also require technical personnel to conduct regular, on-site inspections of power equipment, resulting in considerable expenditure of manpower and material resources. Moreover, inspecting equipment in complex environments or adverse weather conditions poses significant safety risks to personnel \cite{lianqiao2019recognition}. Consequently, these traditional approaches are no longer sufficient to address the needs of modern industrial development and production.

Compared with traditional manual detection methods, deep learning offers significant advantages in object detection, including high efficiency, precision, and cost-effectiveness. Traditional deep learning techniques typically utilize sliding window algorithms to scan target images incrementally, generating multiple candidate regions for further analysis. Feature extraction is then performed on these regions, followed by classification using methods such as support vector machines (SVM) \cite{wang2023fast}. Prominent feature extraction approaches include the Viola-Jones detector \cite{viola2001rapid},  the histogram of oriented gradients (HOG) detector \cite{dalal2005histograms},  and adaboost-based ensemble learning algorithms \cite{freund1997decision}.  However, these methods often depend heavily on expert-designed features, making them susceptible to false detections caused by background noise. Therefore, traditional machine learning-based object detection methods have proven insufficient to meet the demands of modern industrial production.

In recent years, development in artificial intelligence have propelled the capabilities of deep learning, particularly in image processing tasks. Deep learning techniques offer unmatched speed and accuracy, making them increasingly attractive for industrial applications \cite{fahim2024enhancing}.  Consequently, researchers have focused on applying deep learning models to the target detection of power equipment  \cite{yuan2024yolov7}. Among these, You only look once (YOLO) , a real-time object detection algorithm, has gained widespread attention. Unlike traditional methods, YOLO eliminates the need for pre-generated candidate regions, directly predicting the class and location of targets within an image. Since its inception in 2015, YOLO has undergone significant advancements, with the latest version, YOLOv11, demonstrating substantial improvements in detection speed and performance \cite{khanam2024yolov11}.  Given these capabilities, the YOLO deep learning framework holds substantial promise for power equipment object detection, offering a robust solution to the challenges of modern industrial requirements.

\section{Experimental design of power equipment object detection based on YOLO}
YOLO is a state-of-the-art one-stage object detection algorithm renowned for its efficiency and simplicity. The YOLO framework encompasses multiple components, including the construction of a object detection dataset, image preprocessing, model training using the object detection training dataset, and validation of results using a verification dataset. Over successive iterations, YOLO’s backbone network has undergone substantial advancements, integrating deeper feature fusion and multiscale feature extraction to enhance its capability for power equipment object detection. Since YOLOv5 \cite{abdallah2024real},  the algorithm has significantly improved detection efficiency and accuracy through the introduction of the CSPNet framework, which optimizes feature propagation and network capacity. Starting from YOLOv8 \cite{varghese2024yolov8},  the series adopted an anchor-free mechanism for the first time, allowing greater adaptability to detect power equipment targets of varying sizes. Concurrently, updates to the YOLO series have included innovative enhancements to the loss function, further refining the model's detection precision. While the original YOLO algorithm offered remarkable detection speed, its accuracy lagged behind two-stage detection algorithms. However, with iterative updates, newer YOLO versions have achieved substantial improvements in detection accuracy while preserving rapid processing speeds. Notably, the YOLOv11 \cite{zou2024hidden} model represents a significant milestone, achieving a balanced trade-off between speed and accuracy, even surpassing the performance of some two-stage algorithms. This makes YOLOv11 highly effective for power equipment object detection, offering both precision and efficiency essential for modern applications. Fig. \ref{fig1} illustrates the overall architecture of the YOLOv11 network.

\begin{figure}
  \begin{center}
  \includegraphics[width=0.48\textwidth]{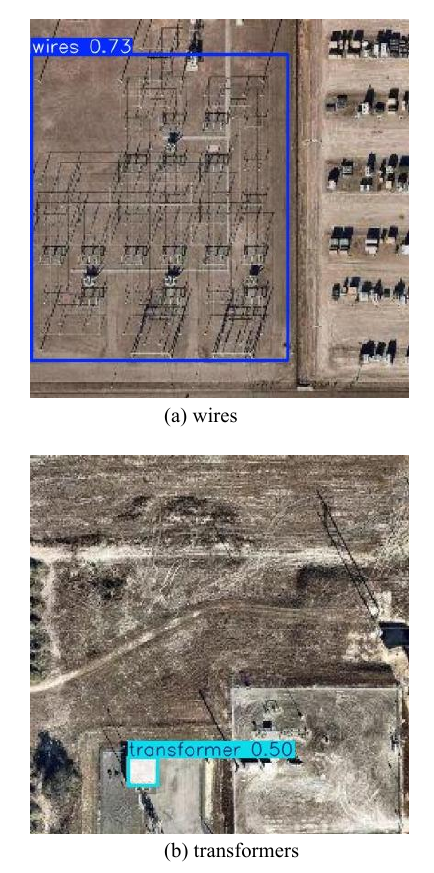}
  \setlength{\abovecaptionskip}{-1pt}
  \caption{The power equipment label of two categories:  wires and transformers.}
  \label{fig2}
  \end{center}
\end{figure}

\subsection{Update of YOLO model}
YOLOv3 \cite{farhadi2018yolov3} introduced multiscale prediction, enabling the detection of bounding boxes at three distinct scales. This enhancement significantly improved the model's capability to detect objects of varying sizes. The incorporation of a spatial pyramid pooling (SPP) layer into the backbone network further expanded the model's receptive field, enhancing its feature extraction capabilities. YOLOv5 advanced these capabilities by adopting the C3 module in its backbone network, which reduced computational complexity and improved inference speed. It also introduced Mosaic data augmentation, particularly Mosaic4, which combines and transforms four images randomly to enhance feature representation and model learning. Adaptive anchor box optimization was added, enabling the model to better handle objects of different sizes.YOLOv8 refined the architecture further by replacing the C3 module with the C2f module, enhancing feature extraction efficiency. It also introduced an Anchor-Free detection mechanism to improve the detection of small targets. The Mosaic augmentation process was optimized to exclude its use in the final ten training epochs, thereby improving model generalization. Additionally, task-specific loss optimizations were integrated to further enhance detection performance. YOLOv9 \cite{wang2402yolov9} introduced progressive gradient integration (PGI), addressing limitations of deep supervision in extremely deep architectures and making lightweight architectures more practical. A new network architecture,  called generalized high-efficiency layer aggregation network (GELAN), was proposed. GELAN integrates cross stage partial network (CSPNet) and efficient layer aggregation network (ELAN) designs, balancing model lightweight design, inference speed, and accuracy. Cross-stage partial connections were employed to link feature maps across stages, enriching semantic information and improving detection accuracy. YOLOv10 \cite{wang2405yolov10} incorporated a dual-head architecture with one-to-many and one-to-one heads. During training, both heads were engaged, while only the One-to-One head was used for inference, enhancing prediction precision. The inclusion of a PSA module after SPPF promoted global feature learning and further improved detection performance.The most recent iteration, YOLOv11, replaced the C2f module with the C3K2 module, a custom CSP bottleneck layer featuring two smaller convolutional layers, improving processing speed without compromising performance. While retaining the SPPF module from YOLOv8, YOLOv11 introduced the C2PSA module, which integrates channel and spatial information with multi-head attention mechanisms for more efficient feature extraction. An adaptive anchor box mechanism was also refined to optimize configurations across diverse datasets, boosting detection accuracy. Beyond object detection in power equipment, YOLOv11 extends its capabilities to instance segmentation, image classification, pose estimation, and oriented bounding box detection (OBB), addressing a wide array of computer vision tasks and challenges.

\subsection{Power equipment object detection method based on YOLOv11}
The YOLOv11 object detection method enhances its performance by minimizing a comprehensive loss function that integrates multiple components. This loss function encompasses distributed focal loss, bounding box regression loss, and class probability loss. The optimization process involves combining these individual loss components and employing advanced optimization algorithms to refine the model's performance in object detection tasks. The specific formulation of the YOLOv11 loss function is expressed by 
\begin{equation}
\begin{array}{l}
{L_{YOLOv11}} = {L_{cls}} + {L_{box}} + {L_{dfl}}
\end{array}
\end{equation}

where $L_{cls}$ is the class probability loss, a pivotal component in the YOLOv11 model's training regimen. It serves as a metric to quantify the divergence between the predicted class probabilities and the ground truth. This loss function, grounded in the principles of cross-entropy, is instrumental in refining the model's predictive accuracy across various object categories. $ L_{box}$ is the boundary box loss, another cornerstone of the YOLOv11 model's optimization strategy, is designed to enhance the precision of bounding box predictions. It achieves this by minimizing the discrepancy between the model's predicted bounding boxes and the actual bounding boxes, leveraging metrics such as the intersection over union (IoU) to quantify the overlap and thus guide the model's learning process. $L_{dfl} $ is the innovative approach of the YOLOv11 model lies in its ability to discern and prioritize challenging samples within the detection landscape. By modulating the loss weights of these complex samples, the model is incentivized to allocate greater focus and computational resources to their accurate classification, thereby bolstering its overall detection prowess.

The YOLOv11 model is structured into three primary components: the Backbone, Neck, and Head. The Backbone serves as the foundation for extracting multi-scale feature maps from input images. This is achieved through a series of convolutional layers and specialized modules, designed to generate feature maps at varying resolutions. These feature maps capture the spatial and semantic information necessary for subsequent processing. The Neck functions as an intermediate stage, tasked with aggregating and enhancing features from multiple scales before passing them to the Head network for prediction. This process often involves upsampling and concatenation of feature maps, enabling the model to efficiently capture and utilize multi-scale information. The Neck plays a crucial role in bridging the Backbone and Head components, enhancing feature expressiveness and supporting robust predictions. The Head is responsible for generating the final outputs, including object bounding boxes and category labels. It processes the enriched feature maps from the Neck to predict object locations and classifications with high precision. 

In summary, the Backbone extracts essential feature representations, the Neck aggregates and refines these features across scales, and the Head produces the final predictions. The Neck serves as a pivotal link, combining multi-scale features from the Backbone and augmenting their expressiveness through upsampling and concatenation, thereby providing a strong foundation for the Head’s accurate and reliable predictions.

\subsection{Experiment Setup}
In this article, YOLOv5, YOLOv8, YOLOv9, YOLOv10, and YOLOv11 were comprehensively evaluated. Each model was trained for 100 epochs with a batch size of 32, employing stochastic gradient descent (SGD) as the optimizer and an initial learning rate of 0.01. The experimental setup included 10 worker threads, an IoU threshold of 0.5, and a standardized input image size of 640 × 640 pixels.The experiments were conducted on a cloud server equipped with an NVIDIA Tesla V100 GPU (16 GB), a 10-core Intel Xeon Platinum 8160T CPU, and 16 GB of RAM. The development environment comprised Python 3.8, CUDA 11.3, and PyTorch 1.10, running on a Windows 11 operating system.

\subsection{Experimental results and analysis}
In this article, a dataset for power equipment object detection was used to evaluate model performance. Specifically, we employed an public power equipment object detection dataset obtained from the Roboflow platform. This dataset includes two labeled object categories: wires and transformer. The defect distribution within the dataset is illustrated in Fig. \ref{fig2}, comprising a total of 497 images. The dataset was partitioned into a training dataset and a validation dataset, with 397 images allocated to the training dataset and 100 images reserved for validation, adhering to a 4:1 ratio.

\begin{figure*}
  \begin{center}
  \includegraphics[width=0.92\textwidth]{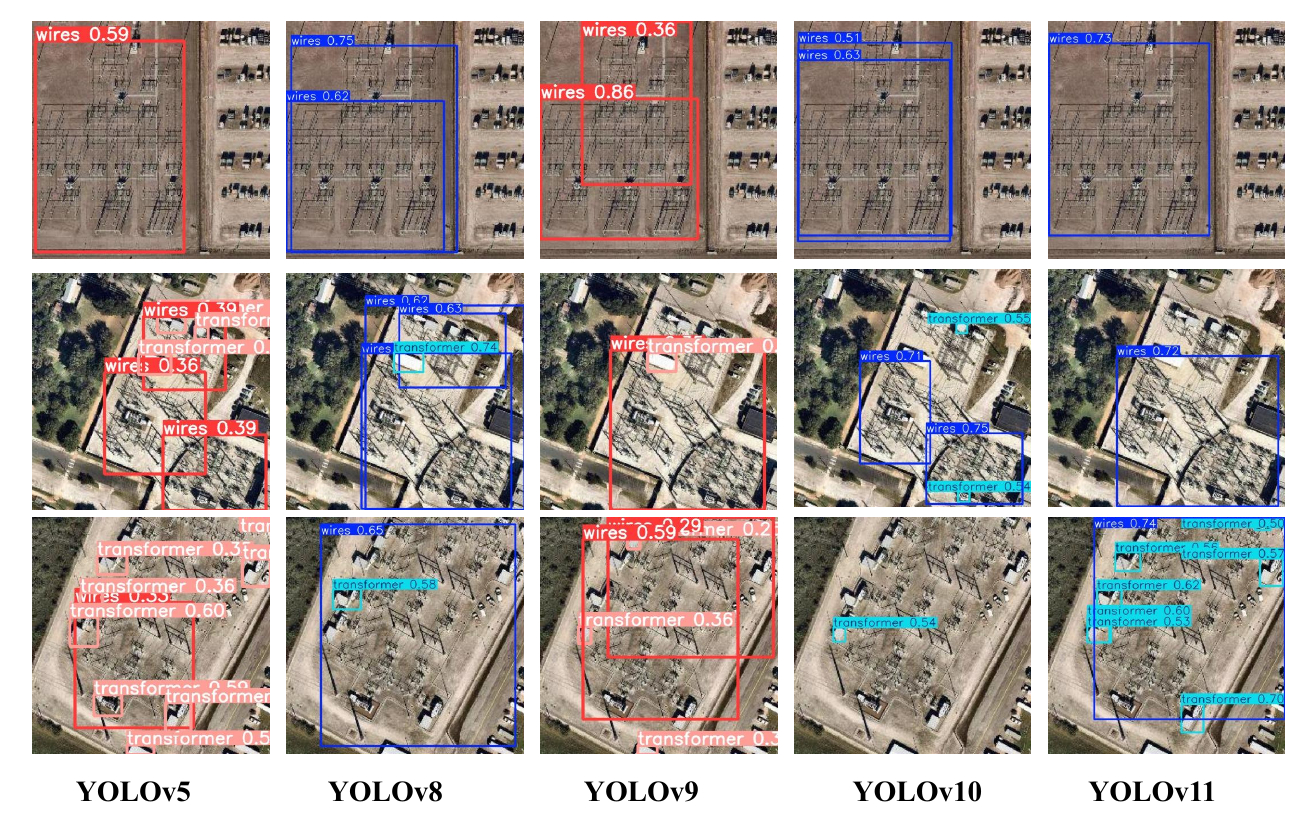}
  \setlength{\abovecaptionskip}{-1pt}
  \caption{The experimental results visualization of YOLO series.}
  \label{fig3}
  \end{center}
\end{figure*}

\begin{figure*}
  \begin{center}
  \includegraphics[width=.92\textwidth]{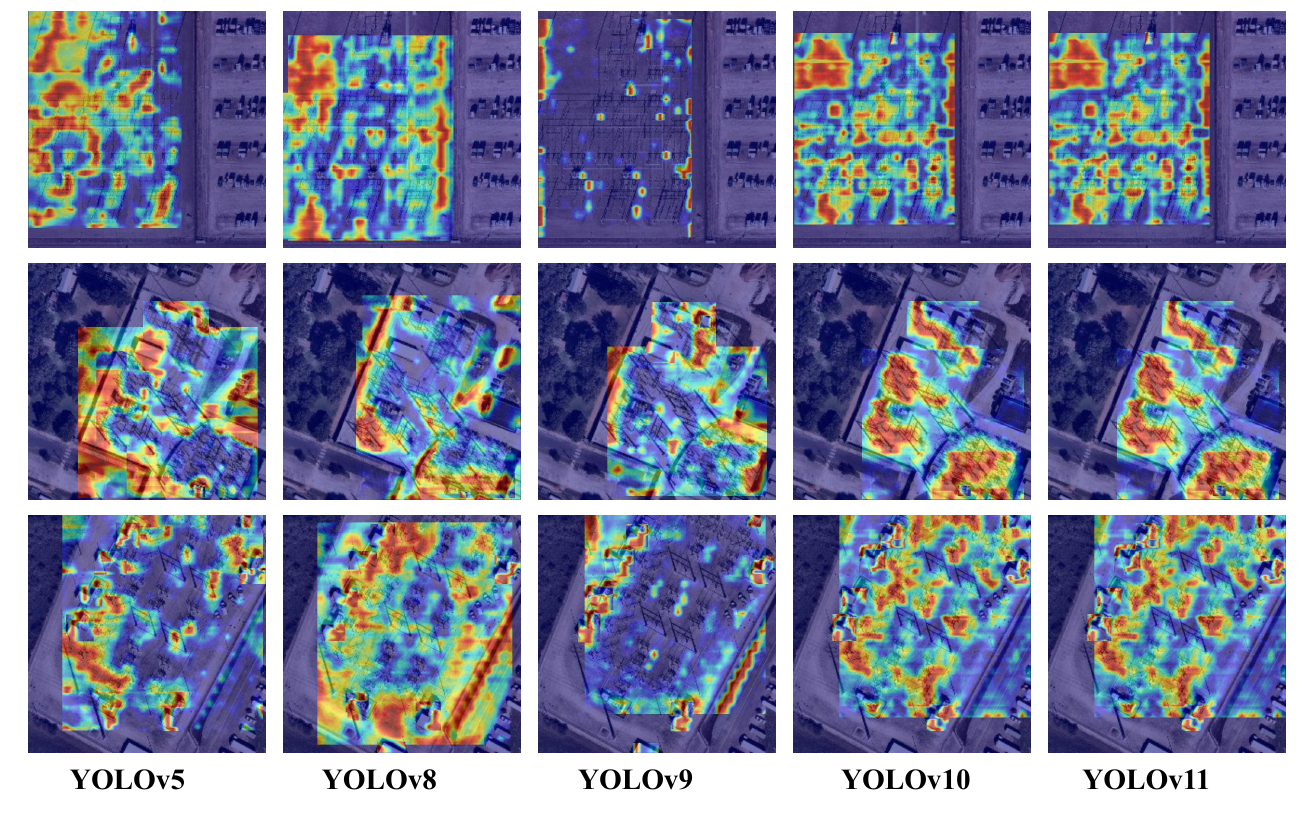}
  \setlength{\abovecaptionskip}{-1pt}
  \caption{ The experiment heat map visualization of YOLO series.}
  \label{fig4}
  \end{center}
\end{figure*}

 The performance results of the tested configurations are presented in Fig. \ref{fig3}. It can be observed that the YOLOv11 model demonstrates significant performance improvements in detection accuracy, missed detections, and false detections for both single and multiple power equipment object scenarios. Therefore, the newly developed YOLOv11 model exhibits substantial application value in power equipment target detection. In order to illustrate the excellent feature extraction ability of YOLOv11 model, Grad-CAM \cite{selvaraju2020grad} tool is used to visualize the detection results of YOLOv5, YOLOv8, YOLOv9, YOLOv10 and YOLOv11 in identifying power equipment object. As shown in Fig. \ref{fig4}, the YOLOv11 model exhibits a pronounced concentration of attention within the spatial confines of the power equipment's object detection zone. This focused attention is in stark contrast to the more dispersed attention maps generated by its predecessors, which fail to pinpoint the precise coordinates of the power equipment's target detection area with the same level of precision. The visualizations underscore the YOLOv11 model's enhanced ability to discern and emphasize the most relevant features for accurate object detection in complex environments.

This article uses standard image detection evaluation indicators, including mean average precision (mAP), Recall and Precision, which is written as
\begin{equation}
\begin{array}{l}
m{\rm A}{\rm P} = \frac{{\sum\limits_{i = 1}^{\rm N} {{\rm A}{{\rm P}_i}} }}{{\rm N}}\\
Precision = \frac{{TP}}{{TP + FP}} = \frac{{TP}}{{all\;detections}}\\
Recall = \frac{{TP}}{{TP + FN}} = \frac{{TP}}{{all\;ground\;truths}}
\end{array}
\end{equation}
In object detection, true positives (TP) represent the number of objects correctly identified by the model. False positives (FP) correspond to instances where the model incorrectly classifies background elements as objects, while false negatives (FN) denote objects that the model fails to detect. The mAP evaluates the overall performance of the model across all categories by averaging the average precision of each class. The Precision measures the proportion of predicted positive samples that are correctly classified, providing insight into the model's propensity for false positives. The Recall quantifies the proportion of actual positive samples that the model successfully identifies, reflecting its ability to minimize false negatives. Additionally, all detections refer to the total number of samples identified as positive by the model, whereas all ground truths encompass the total number of actual positive samples present in the dataset. These metrics collectively provide a comprehensive evaluation of the model's detection performance.

\begin{table}[htbp]
  \centering
  \caption{The experiment results of YOLO series.}
    \begin{tabular}{cccc}
    \toprule
    Method & \multicolumn{1}{l}{mAP/\%} & \multicolumn{1}{l}{Precision/\%} & \multicolumn{1}{l}{Recall/\%} \\
    \hline
    YOLOv5 & 54.4  & 64.5  & 62.6 \\
    YOLOv8 & 55.5  & 71.1  & 60.9 \\
    YOLOv9 & 43.8  & 55.2  & 50.7 \\
    YOLOv10 & 48.0    & 79.3  & 56.2 \\
    YOLOv11 & 57.2  & 66.4  & 64.8 \\
    \bottomrule
    \end{tabular}%
  \label{table1}%
\end{table}%
Table \ref{table1} shows the comparison of mAP, Precision and Recall results of YOLOv5, YOLOv8, YOLOv9, YOLOv10 and YOLOv11. The latest model, YOLOv11, achieves the highest results in both the mAP and Recall, with values of 57.2\% and 64.8\%, respectively. The mAP reflects the model's overall detection performance, while the Recall highlights its ability to minimize false negatives. These results demonstrate that the YOLOv11 model exhibits superior detection accuracy and a reduced rate of false detections, making it highly effective for power equipment target detection and showcasing its substantial application potential.

Table \ref{table2} shows the  comparison of the test results for YOLOv5, YOLOv8, YOLOv9, YOLOv10, and YOLOv11 on wires and transformers, using mAP as the evaluation metric. The recently developed YOLOv11 model emerges as the paragon of detection efficacy, demonstrating unparalleled performance across both categorical domains. This preeminent achievement in the realm of power equipment target detection within the experimental framework of this study suggests that YOLOv11 is not merely a incremental advancement but a transformative leap in object detection technology. The implications of these findings are profound, heralding a new era of possibilities for YOLOv11 in the realms of scientific inquiry and practical application scenarios.

\begin{table}[htbp]
  \centering
  \caption{The object detection experiment results of YOLO series in power equipment using mAP (\%).}

    \begin{tabular}{lrrr}
    \toprule
     Method & \multicolumn{1}{l}{wires} & \multicolumn{1}{l}{transformer}  \\
    \toprule
YOLOv5 & 64.8  & 44.7 \\
YOLOv8 & 64.8  & 46.1 \\
YOLOv9 & 53.7  & 33.8 \\
YOLOv10 & 59.0    & 36.9 \\
YOLOv11 & 73.9  & 62.0 \\
   \bottomrule
    \end{tabular}%
  \label{table2}%
\end{table}%
\section{Conclusion}
In this article, YOLOv5, YOLOv8, YOLOv9, YOLOv10, and YOLOv11 are systematically evaluated for object detection of power equipment. In the YOLOv11 model, the C2f module was replaced with the C3K2 module, a custom implementation of the CSP bottleneck layer that employs two smaller convolutional layers instead of the single large convolutional layer used in YOLOv8. This modification not only preserves performance but also enhances processing speed due to the smaller convolution kernel size. The SPPF module from YOLOv8 was retained, and the C2PSA module was added afterward. The C2PSA module integrates channel and spatial information, enabling more efficient feature extraction and working synergistically with multi-head attention mechanisms to improve detection performance. 
Additionally, the adaptive anchor frame mechanism allows the model to automatically optimize anchor configurations across different datasets, further boosting detection accuracy. The experimental results demonstrate that YOLOv11 achieves the highest average accuracy and recall rate in power equipment object detection. These findings underscore its superior detection performance, making YOLOv11 a model with significant application potential and promising future prospects.


\bibliographystyle{unsrt}
\bibliography{mybibfile}

\end{document}